\documentclass[runningheads]{llncs}
\usepackage[T1]{fontenc}
\usepackage{amsfonts} 
\usepackage{amsmath}
\usepackage{graphicx}

\usepackage{graphicx,verbatim}
\usepackage{booktabs}
\usepackage{multirow}

\usepackage{orcidlink}

\begin{document}
\title{From $\mathcal{O}(n^{2})$ to $\mathcal{O}(n)$ Parameters: Quantum Self-Attention in Vision Transformers for Biomedical Image Classification}
\author{Thomas Boucher\inst{1,3}\orcidlink{0009-0002-6152-7271} \and
John Whittle\inst{2,3}\orcidlink{0000-0002-3859-679X} \and
Evangelos B. Mazomenos\inst{1}\orcidlink{0000-0003-0357-5996}}
\authorrunning{T. Boucher \textit{et al.}}
\institute{UCL Hawkes Institute, Department of Medical Physics and Biomedical Engineering, University College London, London, United Kingdom. \and Department of Anaesthesia and Peri-operative Medicine, University College London Hospitals NHS Foundation Trust, London, UK.\and Human Physiology and Performance Laboratory (HPPL), Centre for Peri-operative Medicine, Department of Targeted Intervention, Division
of Surgery and Interventional Science, University College London, London, UK.\\ \email{\{thomas.boucher.23, e.mazomenos\}@ucl.ac.uk}}
\maketitle              % typeset the header of the contribution
\begin{abstract}
We demonstrate that quantum vision transformers (QViTs), vision transformers (ViTs) with self-attention (SA) mechanisms replaced by quantum self-attention (QSA) mechanisms, can match state-of-the-art (SOTA) biomedical image classifiers while using 99.99\% fewer parameters. QSAs are produced by replacing linear SA layers with parameterised quantum neural networks (QNNs), producing a QSA mechanism and reducing parameter scaling from $\mathcal{O}(n^2)$ to $\mathcal{O}(n)$. On RetinaMNIST, our ultra parameter-efficient QViT outperforms 13/14 SOTA methods including CNNs and ViTs, achieving 56.5\% accuracy, just 0.88\% below the top MedMamba model while using 99.99\% fewer parameters (1K vs 14.5M) and 89\% fewer GFLOPs. We present the first investigation of knowledge distillation (KD) from classical to quantum vision transformers in biomedical image classification, showing that QViTs maintain comparable performance to classical ViTs across eight diverse datasets spanning multiple modalities, with improved QSA parameter-efficiency. Our higher-qubit architecture benefitted more from KD pre-training, suggesting a scaling relationship between QSA parameters and KD effectiveness. These findings establish QSA as a practical architectural choice toward parameter-efficient biomedical image analysis.
\footnote{Code available at https://github.com/surgical-vision/QViT-KD.git.}
% Authors must provide keywords and are not allowed to remove this Keyword section.

\end{abstract}
\section{Introduction}
Vision transformers (ViTs) \cite{dosovitskiy2021imageworth16x16words} have emerged as a transformative architecture in computer vision. By treating images as sequences of patches, and leveraging self-attention (SA) to capture long-range dependencies and intricate patterns within image data, ViTs have demonstrated exceptional performance for biomedical classification tasks, often surpassing convolutional neural networks (CNNs) \cite{manzari2023medvit}. However, their computational demands, with $\mathcal{O}(n^2)$ parameter scaling in SA layers, limit deployment in resource-constrained clinical settings.

Quantum machine learning (QML) offers a fundamentally different approach to the efficiency challenge. By leveraging quantum neural networks (QNNs) that exploit quantum mechanical phenomena within complex Hilbert spaces, QML achieves rich, parameter-efficient feature representations unattainable in classical Euclidean spaces. QML has recently witnessed significant advancements, with a surge of applications in biomedical image analysis \cite{wei2023quantum}, where the enhanced representational power enables QNNs to capture intricate data patterns with dramatically improved parameter efficiency, leading to more effective machine learning, particularly for supervised tasks \cite{havlivcek2019supervised}.

Quantum vision transformers (QViTs) merge ViTs and QML by replacing the parameter-heavy linear projections in ViT's SA mechanisms with QNNs, creating quantum self-attention (QSA) mechanisms. This architectural 
 substitution reduces the SA mechanism's parameter scaling from $\mathcal{O}(n^2)$ to $\mathcal{O}(n)$ while maintaining representational power within the expressive complex Hilbert space.
Whilst research into QViTs is still in its nascent stages, recent work has shown these architectures can compete with classical models despite using vastly fewer parameters \cite{cherrat2024quantum}.
\begin{figure}[b!]
\includegraphics[width=\textwidth]{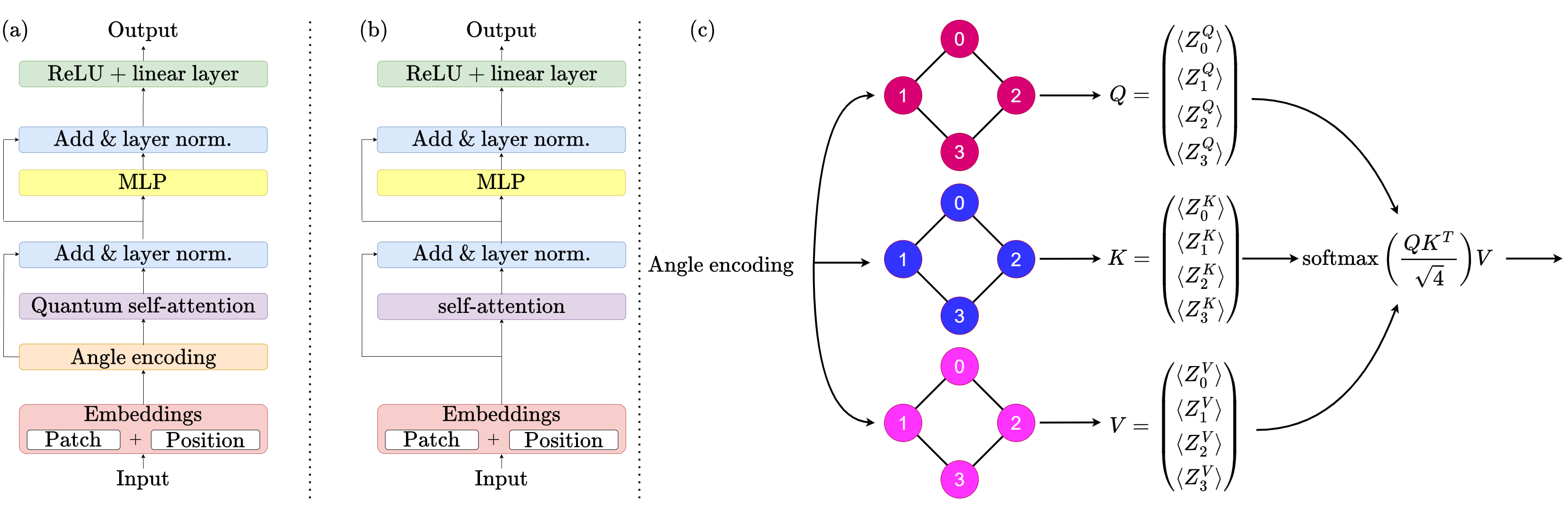}
\caption{(a) The QViT architecture, utilising QSA following angle-encoding in place of classical SA, (b) the respective ViT architecture, (c) a visualisation of the QSA with angle encoding and our chosen ansatz.}
\label{fig:qsa}
\end{figure}

Knowledge distillation (KD), which transfers rich knowledge from large teacher models to efficient students \cite{hinton2015distillingknowledgeneuralnetwork}, offers a compelling direction to further improve QViT performance, that is unexplored in the literature.  
While KD has proven effective for enhancing classical ViTs \cite{wu2022tinyvit}, its application to enhancing QViTs remains uninvestigated. This gap is particularly significant in biomedical imaging, where model efficiency is crucial for potential clinical deployment.
We systematically compare the performance of QViTs against classical ViTs across eight datasets
encompassing distinct imaging modalities, a variety of binary and multi-class classification tasks, and an ordinal regression task. We evaluate the performance of models trained both from scratch and with KD pre-training from a high-quality classical teacher model, under a range of training conditions.

We make the following contributions: 
(1) we present the first investigation into the efficacy of pre-training parameter-efficient QViTs with KD for biomedical image classification, revealing that effectiveness scales with parameters in the QSA; (2) we demonstrate that QViTs compete with ViTs of equivalent parameter number (Fig.~\ref{fig:qsa}(a), (b)) both with and without KD pre-training across eight diverse datasets spanning multiple modalities and classification tasks, whilst utilising more parameter-efficient QSA; (3) we show that extremely parameter-efficient QViTs are capable of outperforming state-of-the-art (SOTA) classical models, achieving near-best accuracy on RetinaMNIST with a 99.99\% reduction in parameters compared to MedMamba, illustrating the potential to produce high-performance, parameter-efficient biomedical image classifiers by replacing SA with more parameter-efficient QSA (Fig.~\ref{fig:qsa}(c)), and through combining QViTs with KD for effective QML.
\section{Primer: Quantum Networks}
\begin{figure}[b!] 
\includegraphics[width=\textwidth]{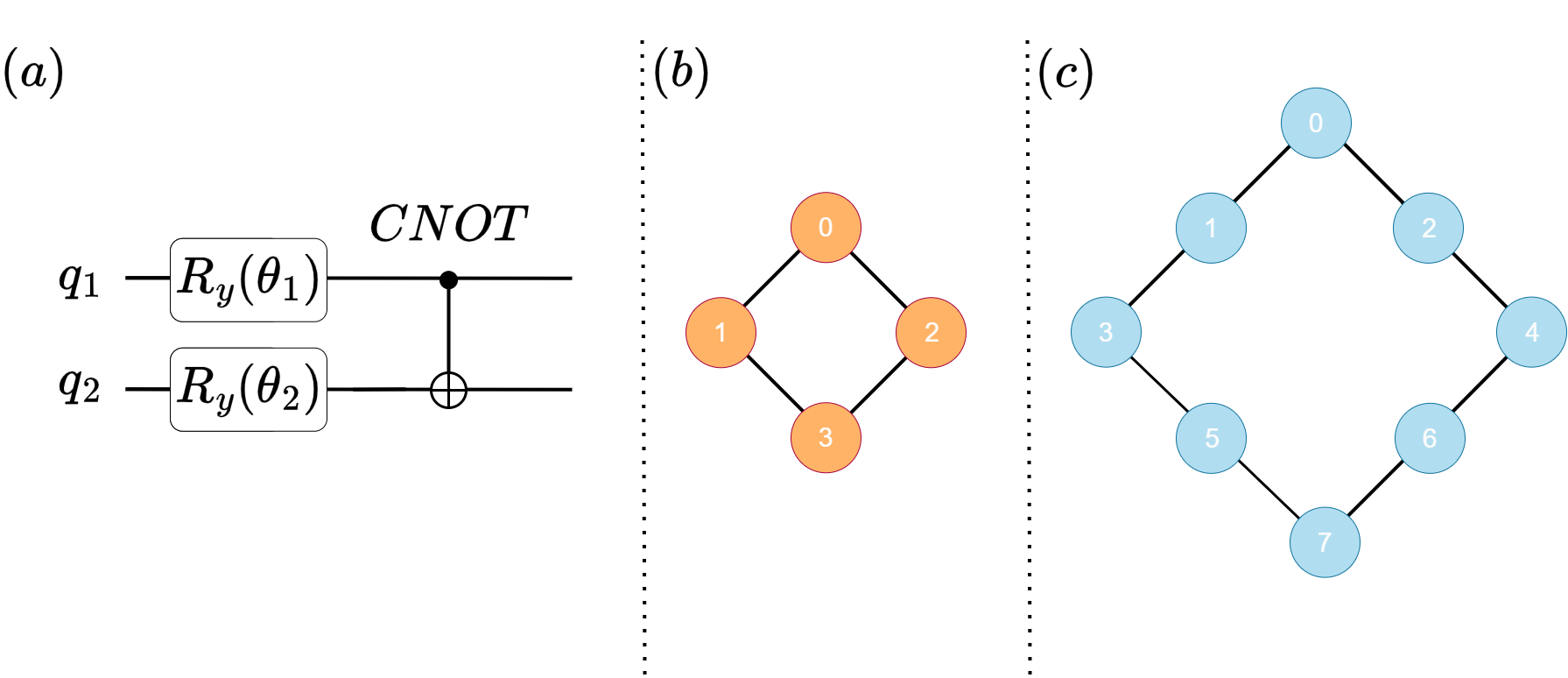}
\caption{(a) The ansatz we use, applied to qubit pairs ($q_{1}$, $q_{2})$, (b) the four-qubit and (c) the eight qubit QNNs structure we use, with paired qubits connected by lines.
}
\label{fig:qnns}
\end{figure}
Quantum networks are comprised of qubits, which represent a continuous superposition of quantum states. These states are parametrised by complex probability amplitudes of unit norm:
\begin{equation}\label{eq:qstate}
    | \psi \rangle = \alpha |0 \rangle + \beta |1 \rangle = \begin{bmatrix}
        \alpha \\ \beta
    \end{bmatrix} \in \mathbb{H}, \quad \alpha, \beta \in \mathbb{C},\quad  |\alpha|^{2} +|\beta|^{2} = 1,
\end{equation}
where $\mathbb{H}$ is a Hilbert space.
This allows qubits to occupy an infinite number of states, enhancing their representational power compared to a bit, which is binary.
\noindent Quantum states are transformed by unitary operations, which are also essential for encoding classical data onto quantum networks via a quantum feature map $x \in \mathbb{R} \to |\psi( x) \rangle \in \mathbb{H}$. 
For an $n$-qubit network initialised in a state $| 0 \rangle^{n} = \bigotimes^{n}_{i=1} |0 \rangle,$ this encoding is a unitary transformation $U_{\psi}(x)|0\rangle^{n} = | \psi(x) \rangle$. We use angle encoding, which encodes $n$ classical bits onto $n$ qubits. For $x \in \mathbb{R}^{n}$:
\begin{equation}\label{eq:qfm}
    U_{\psi} = \bigotimes^{n}_{i=1} U_{x_{i}}, \quad U_{x_{i}} := \left[ {\begin{array}{cc}
        cos(x_{i}/2) & -sin(x_{i}/2)\\
        sin(x_{i}/2) & cos(x_{i}/2)\\
    \end{array} } \right]
\end{equation} 
\noindent This is chosen for its efficiency and high performance in quantum classifiers \cite{grant2018hierarchical}. This encoded state can then be transformed by further parametrised unitary operations $U(\Theta)|\psi(x) \rangle = |\psi'(x) \rangle$, $U(\Theta) \in \text{U}(n)$. 
A specific sequence of parametrised unitary operations is an ansatz, which is typically designed to act on qubit pairs. We use a particularly parameter-efficient ansatz, visualised in Fig.~\ref{fig:qnns}(a), that is shown to be effective with angle encoding for classification tasks \cite{hur2022quantum}. This ansatz applies a parametrised unitary operation $R_{y}(\theta_{i})$ to each qubit for $i=1,2$, followed by a controlled-NOT (CNOT):
\begin{equation} \label{eq:ry_cnot}
    R_{y}(\theta_{i}) := \left[ {\begin{array}{cc}
    cos(\theta_{i}/2) & -sin(\theta_{i}/2)\\
    sin(\theta_{i}/2) & cos(\theta_{i}/2)\\
  \end{array} } \right], \quad \text{CNOT} := \left[ {\begin{array}{cccc}
        1\;\; & 0\;\; & 0\;\; & 0\\
        0\;\; & 1\;\; & 0\;\; & 0\\
        0\;\; & 0\;\; & 0\;\; & 1\\
        0\;\; & 0\;\; & 1\;\; & 0\\
   \end{array} } \right],
\end{equation} 
\noindent where CNOT, defined by $\text{CNOT}\left(|q_{1}\rangle \otimes |q_{2} \rangle\right):= |q_{1}\rangle \otimes |(q_{1} \oplus q_{2})\bmod{2} \rangle $, entangles the two qubits, creating a joint state which is essential for quantum algorithms to achieve computational advantages over classical methods \cite{jozsa2003role}.

To extract classical information from the network, a unitary operation $Z$ is used to measure the expectation value $\langle Z \rangle $ of each qubit, defined by:
\begin{equation}
    \langle Z \rangle := \langle \psi | Z | \psi \rangle = \begin{bmatrix}
        \alpha &
        \beta
    \end{bmatrix}
    \begin{bmatrix}
        1 & 0\\
        0 & -1
    \end{bmatrix}
    \begin{bmatrix}
        \alpha \\ \beta
    \end{bmatrix} = |\alpha|^{2} - |\beta|^{2} \in \mathbb{R}.
\end{equation}\label{eq:expect}
\noindent Performing this measurement for each qubit produces a vector $y \in \mathbb{R}^{n}$.
\section{Methods}

\subsection{Architecture Design}
We use the QSA architecture of \cite{cara2024quantum}, where QNNs replace the linear projection layers for key, query, and value in the SA to produce a QSA mechanism. In this study, $n$-qubit QNNs (Fig.~\ref{fig:qnns}(b, c)) replace $n\to n$ linear projections, and the resulting QViTs are compared to the original ViTs. While both maintain $\mathcal{O}(n^{2})$ computational complexity, they differ fundamentally in parameter scaling: QSAs require only $\mathcal{O}(n)$ parameters ($6n$ with our ansatz) vs $\mathcal{O}(n^{2})$ for ViTs (specifically $3n^2$). This comparison methodology follows standard practice for assessing quantum advantage for low qubit configurations \cite{herrmann2023quantum}.

For both architectures, patch embeddings are extracted using convolutions and combined with positional embeddings. These embeddings are processed by the SA mechanism, followed by a multilayer perceptron, before passing through a rectified linear unit (ReLU) and final linear layer for classification.

\begin{figure}[t!]
\includegraphics[width=\textwidth]{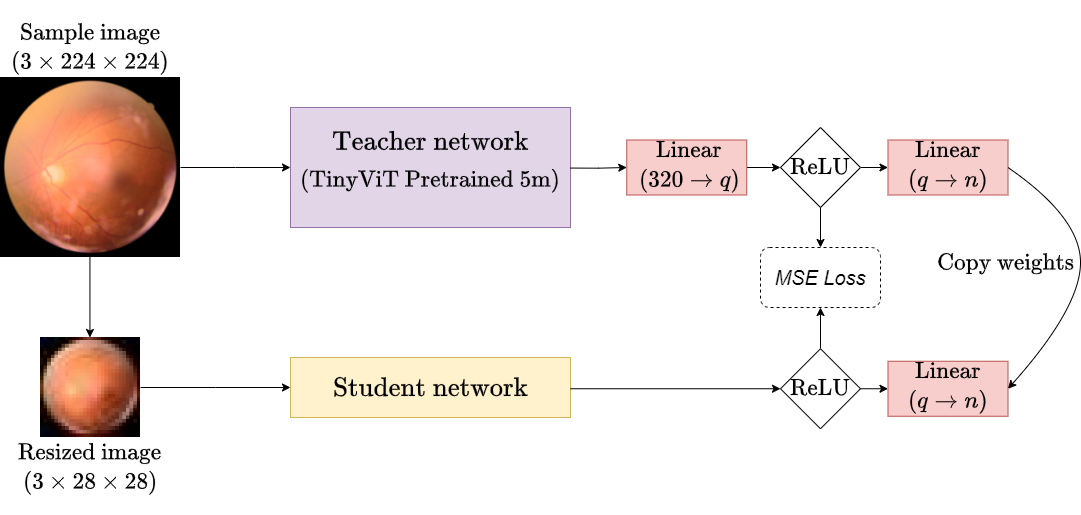}
\caption{The KD architecture for a sample image from RetinaMNIST with a $28\times 28$ input student network, where $q$ is the number of qubits, $n$ is number of target classes. The student is optimised to minimise the MSE loss between the teacher's and student's intermediate logits. Following KD training, the teacher's final linear layer weights are copied to the student's final layer. For $224 \times 224$ inputs, the resizing step is skipped.}
\label{fig:kd}
\end{figure}

\subsection{Knowledge Distillation Framework}
%We present the first classical-to-quantum KD framework for biomedical imaging. 
We use the ImageNet-pre-trained TinyViT model \cite{wu2022tinyvit} (5M parameters) as our teacher network, fine-tuning it independently on each dataset to create task-specific teachers. Its moderate size makes it suitable for distilling to extremely parameter-efficient students \cite{cho2019efficacy}.

To enable effective knowledge transfer between classical and quantum models, we modified TinyViT's classification head by replacing the original linear layer with two linear layers: the first mapping to the number of qubits ($n$), and the second mapping to the number of output classes, with a ReLU activation function in-between. This intermediate layer provides compatible targets for distillation, as the quantum student naturally produces $n$-dimensional outputs from its $n$-qubit measurements. During KD pre-training, students learn to match the teacher's intermediate representations over 50 epochs with Adam optimiser (learning rate $1 \times 10^{-3}$, batch size 32), and using MSE loss, which has been shown to be most effective for KD \cite{kim2021comparing}.

Following KD, we transfer the teacher's final linear layer weights directly to the student, leveraging the aligned intermediate representations. This approach outperformed direct logit distillation, improving accuracy by 9.31\% for ViT and 14.1\% for QViT. The complete KD architecture is visualised in Fig.~\ref{fig:kd}.

\subsection{Experimental Setup}
We evaluate three configurations evaluating 4-qubit models on $28\times 28$ images (2×2 patches) and $224\times 224$ images (16×16 patches), and 8-qubit models on $224\times 224$ images (16×16 patches).
The specific qubit topologies (Fig.~\ref{fig:qnns} (b, c)) match IBM and Rigetti quantum processor connectivity, ensuring our simulated results translate to real hardware.

\noindent\textbf{Training Protocol:} Following \cite{yang2023medmnist}, we use Adam optimiser with initial learning rate $1 \times 10^{-3}$, decaying by $10\times$ at epochs 50 and 75, batch size 32, and 100 total epochs. Models are evaluated both trained from scratch (QViT/ViT) and with 50 epochs of KD pre-training before fine-tuning (QViT-KD/ViT-KD).

\noindent\textbf{Datasets:} We evaluate across eight biomedical imaging tasks (Table~\ref{tab:datasets}): seven from MedMNIST \cite{yang2023medmnist} spanning multiple modalities and classification types with predefined train/validation/test sets, plus OASIS \cite{marcus2007open} for Alzheimer's progression classification. For OASIS, axial slices were used as input, with only slices between the 100th and 160th slice considered for each patient. We used a subset of the full dataset for a better balance between the four classes and for time efficiency. Our dataset selection ensures robust evaluation across binary and multi-class classification, and an ordinal regression task (RetinaMNIST).

\noindent\textbf{Implementation:} All experiments utilised PennyLane \cite{bergholm2018pennylane} with PyTorch for quantum circuit simulation on a single NVIDIA GeForce RTX 4090 GPU.

\begin{table}[t!]
    \centering
    \caption{The details of the eight datasets used in this research.}
    \label{tab:datasets}
    \begin{tabular}{@{}p{0.25\textwidth}p{0.28\textwidth}p{0.15\textwidth}p{0.3\textwidth}@{}}
        \toprule
        \textbf{Name} & \textbf{Modality} & \textbf{Classes} & \textbf{Train/Val/Test}\\
        \midrule
        BreastMNIST & Breast Ultrasound & 2  & 546 / 78 / 156\\
        RetinaMNIST & Fundus Camera & 5  & 1,080 / 120 / 400\\
        PneumoniaMNIST & Chest X-Ray & 2  & 4,708 / 524 / 624\\
        DermaMNIST & Dermatoscope & 7  & 7,007 / 1,003 / 2,005 \\
        BloodMNIST & Blood Cell Microscope & 8  & 11,959 / 1,712 / 3,421\\
        OrganCMNIST & Abdominal CT & 11  & 13,000 / 2,392 / 8,268\\
        PathMNIST & Colon Pathology & 9  & 89,996 / 10,004 / 7,180\\
        OASIS & Brain MRI & 4 & 5320 / 896 / 512 \\
        \bottomrule
    \end{tabular}
\end{table}

\section{Results}

\begin{table}[t!]
    \centering
    \caption{Performance comparison of QViT and ViT. Top score in each metric for each dataset is in bold.}
    \label{tab:combined_results}
    \resizebox{\textwidth}{!}{%  Scale to the width of the text
    \begin{tabular}{@{}l|cc|cc|cc|cc@{}}
        \toprule
        \multirow{2}{*}{\textbf{Methods}} & \multicolumn{2}{c|}{\textbf{BreastMNIST}} & \multicolumn{2}{c|}{\textbf{RetinaMNIST}} & \multicolumn{2}{c|}{\textbf{PneumoniaMNIST}} & \multicolumn{2}{c}{\textbf{DermaMNIST}} \\
        & AUC & ACC & AUC & ACC & AUC & ACC & AUC & ACC  \\
        \midrule
        \multicolumn{9}{c}{\textbf{Trained from Scratch}} \\
        \midrule
        ViT\_28 &0.733 &\textbf{0.821}	&0.690	&0.468	&0.931	&0.846	&0.863	&0.694\\
        ViT\_224 & 0.709& 0.744& 0.696& 0.528& 0.931& 0.849& 0.870& 0.691\\
        ViT\_224 (8 qubits) & 0.749 & 0.814 & 0.705 & 0.510 & 0.949 & 0.830 & \textbf{0.900} & \textbf{0.729}\\
        QViT\_28 & \textbf{0.788}& 0.801& \textbf{0.748}& \textbf{0.565}& 0.904& \textbf{0.862}& 0.857& 0.695\\
        QViT\_224 & 0.735 &0.763 & 0.683 & 0.525 & \textbf{0.944} & 0.856 & 0.862 & 0.696 \\
        QViT\_224 (8 qubits) & 0.675 & 0.731 & 0.722 & 0.523 & 0.940 & 0.827 & 0.883 & 0.702\\
        \midrule
        \multicolumn{9}{c}{\textbf{Knowledge Distillation Pretraining}} \\
        \midrule
        ViT-KD\_28  & 0.770& 0.756& 0.701& 0.518 &0.935& 0.824& 0.865 &0.696\\
        ViT-KD\_224  & 0.681& 0.731& 0.708& 0.488& 0.938& 0.825& 0.857& 0.685\\
        ViT-KD\_224 (8 qubits)  & 0.647 & 0.731 & 0.699 & 0.500 & 0.937 & 0.814 & \textbf{0.896} & \textbf{0.725}\\
        QViT-KD\_28 &0.636& 0.750& \textbf{0.744}& \textbf{0.535}& 0.938& \textbf{0.864} & 0.849& 0.693\\
        QViT-KD\_224 & 0.714	& 0.763 & 0.720 & 0.510 & \textbf{0.943} & 0.840 & 0.860 & 0.701\\
        QViT-KD\_224 (8 qubits)  & \textbf{0.786} & \textbf{0.808} & 0.722 & 0.515 & 0.933 & 0.830 & 0.890 & 0.719\\
        \midrule
        TinyViT (Teacher) & 0.900 & 0.904 & 0.790 & 0.618 & 0.988 & 0.886 & 0.933 & 0.823  \\
        \midrule
        \midrule
        \multirow{2}{*}{\textbf{Methods}} & \multicolumn{2}{c|}{\textbf{BloodMNIST}} &  \multicolumn{2}{c|}{\textbf{OrganCMNIST}} & \multicolumn{2}{c|}{\textbf{PathMNIST}} & \multicolumn{2}{c}{\textbf{OASIS}} \\
        & AUC & ACC & AUC & ACC & AUC & ACC & AUC & ACC \\
        \midrule
        \multicolumn{9}{c}{\textbf{Trained from Scratch}} \\
        \midrule
        ViT\_28 & 0.979&	0.844&	0.912&	0.574&	0.932&	0.686 &  \textbf{0.822} & \textbf{0.701}\\
        ViT\_224 & 0.976& 0.835& 0.938& 0.627 & 0.944 & 0.687 & 0.603 & 0.400 \\
        ViT\_224 (8 qubits) & \textbf{0.996} & \textbf{0.944} & \textbf{0.977} & \textbf{0.792} & \textbf{0.970} & \textbf{0.813} & 0.785 & 0.684\\
        QViT\_28 & 0.968 & 0.812 & 0.923 & 0.618 & 0.934 & 0.684 & 0.812 & 0.693\\
        QViT\_224 & 0.969 & 0.815 & 0.947 & 0.633 & 0.933 & 0.650 & 0.785 & 0.678\\
        QViT\_224 (8 qubits) & 0.972 & 0.816 & 0.965 & 0.719 & 0.930 & 0.762 & 0.767 & 0.682\\
        \midrule
        \multicolumn{9}{c}{\textbf{Knowledge Distillation Pretraining}} \\
        \midrule
        ViT-KD\_28  & 0.976& 0.839& 0.914& 0.613& 0.942& 0.709 & 0.810 & 0.693 \\
        ViT-KD\_224  & 0.981& 0.863& 0.906& 0.555 & 0.942& 0.682 & 0.749 & 0.652\\
        ViT-KD\_224 (8 qubits)  & \textbf{0.995} & \textbf{0.939} & \textbf{0.978} & \textbf{0.798} & 0.959 & 0.789 & \textbf{0.824} & \textbf{0.697}\\
        QViT-KD\_28 & 0.972 & 0.824 & 0.928 & 0.612 &  0.921 & 0.660 & 0.764 & 0.641\\
        QViT-KD\_224 & 0.971 & 0.815 & 0.917 & 0.613 & 0.939 & 0.712 & 0.763 & 0.656 \\
        QViT-KD\_224 (8 qubits)  & \textbf{0.995} & 0.937 & 0.974 & 0.788 & \textbf{0.970} & \textbf{0.822} & 0.777 & 0.684\\
        \midrule
        TinyViT (Teacher) & 0.999 & 0.989 & 0.997 & 0.956 & 0.990 & 0.889 & 0.983 & 0.965\\
        \bottomrule
    \end{tabular}}
\end{table}

Our primary findings, detailed in Table~\ref{tab:combined_results}, demonstrate that replacing SA with QSA can achieve high performance with improved parameter efficiency.

\noindent\textbf{QViTs are Competitive with Classical ViTs:} When trained from scratch, QViTs consistently perform on par with their classical ViT counterparts that share the same structural design but have quadratically more parameters in their SA mechanism. For instance, across the four-qubit configurations (ViT\_28 vs. QViT\_28 and ViT\_224 vs. QViT\_224), the quantum models achieve comparable or slightly better average performance (48 SA parameters vs. 24 QSA parameters). This trend continues in the eight-qubit setting, where the ViT model's larger parameter count in its attention block (192 SA parameters vs. 48 QSA parameters) does not translate to a consistent performance advantage.

\noindent\textbf{SOTA-Level Performance with Extreme Parameter Efficiency:} The most striking result is the performance of our 4-qubit QViT\_28 on the RetinaMNIST dataset. This model, with only 1K total parameters, achieved 56.5\% accuracy, outperforming 13/14 SOTA classifiers, including various ResNets, ViTs, and all but one MedMamba variant \cite{yue2024medmamba}. This near-SOTA performance was achieved with just a 0.88\% accuracy gap to the top MedMamba model, while using 99.99\% fewer parameters (1K vs. 14.5M) and requiring 89\% fewer GFLOPs. This result powerfully illustrates the potential of QSA to contribute to SOTA models with significant parameter-efficiency.

\noindent\textbf{Knowledge Distillation Effectiveness Scales with Quantum Capacity:} Our investigation into classical-to-quantum KD reveals a nuanced relationship between QSA capacity and KD effectiveness. For the low-parameter QSA 4-qubit models, KD pre-training did not yield a performance benefit and in some cases was detrimental compared to training from scratch. However, for the 8-qubit QViT, which has double the quantum parameters in its QSA mechanism, KD pre-training provided a clear performance boost, improving average accuracy from $0.720$ to $0.748$ and bringing its performance in line with its $8$-qubit ViT counterpart. This suggests that a minimum level of parametric capacity in the QSA mechanism is necessary for the student QViT to effectively absorb the distilled knowledge from a complex classical teacher.

\section{Discussion}
Our research establishes QViTs as a compelling architecture for biomedical image classification, demonstrating that replacing classical SA with QSA in ViTs can yield models that are competitive with SOTA methods while being orders of magnitude more parameter-efficient.

The central insight from our work is the relationship between a QSA's parametric capacity and its ability to benefit from KD. $4$-qubit models failed to benefit from KD, with QViT-KD\_28 performing 2.65\% and 2.64\% worse than its scratch-trained counterpart for ROC AUC and accuracy, and QViT-KD\_224 performing marginally worse also. This aligns with classical findings that student models require sufficient capacity to learn from teachers \cite{cho2019efficacy}. However, eight-qubit QViTs showed marked improvement with KD, achieving $0.876$ and $0.748$ average AUC and accuracy compared to $0.857$ and $0.720$ without pre-training. The successful application of KD on our $8$-qubit QViT provides the first evidence in this domain that as the capacity of the QSA mechanism increases, so does its potential to leverage pre-training from larger classical models. As developing quantum hardware enables higher stable qubit counts, this relationship suggests a clear scaling path, where larger QViTs combined with KD should yield increasingly competitive performance while maintaining parameter efficiency.

A current and practical limitation of our work is the reliance on classical simulation of quantum circuits, where computational cost scales exponentially ($\mathcal{O}(2^n)$) with the number of qubits $n$. This restricted our experiments to $4$- and $8$-qubit QNNs, which naturally limits the maximal performance achievable. However, this represents an engineering bottleneck rather than a fundamental limitation. The distributed architecture of the QSA is well-suited for near-term networked quantum processors \cite{main2025distributed}, and breakthroughs in fault-tolerant quantum computing \cite{acharya2024quantum} promise stable 100+ qubit systems, and facilitating the deployment of more powerful QViTs to clinical settings in the fear future.

Our work provides a strong proof-of-concept for the utility of QSAs within ViTs for biomedical image analysis. By achieving near-SOTA performance on RetinaMNIST with a 99.99\% parameter reduction, and demonstrating that KD can benefit QViTs with appropriate QSA parameter number, we have shown that QSA offers a practical path toward developing ultra-efficient yet powerful hybrid classical-quantum models. This is particularly relevant for deployment in resource-constrained clinical settings where computational efficiency is paramount. Our results lay the groundwork for developing the next generation of hybrid quantum-classical models for SOTA, parameter-efficient biomedical image analysis.

\section*{Acknowledgments}
This work was supported by the EPSRC-funded UCL Centre for Doctoral Training in Intelligent, Integrated Imaging in Healthcare (i4health) [EP/S021930/1]; the International Anesthesia Research Society Mentored Research grant, and the NIHR University College London Hospitals Biomedical Research Centre.
\bibliography{ForArXiV}

\end{document}